\documentclass[letterpaper,twocolumn,fleqn]{article} 

\usepackage{ist}
\usepackage{textcomp}
\usepackage{gensymb}
\usepackage{multirow}
\usepackage{times}
\usepackage{setspace}
\title{LiDAR-assisted Large-scale Privacy Protection in Street-view \\ Cycloramas}

\author{Clint Sebastian, Bas Boom, Egor Bondarev, Peter H.N. de With; \\ 
Eindhoven University of Technology, Eindhoven, The Netherlands; \\
Cyclomedia B.V, Zaltbommel, The Netherlands}

\date{} 

\begin{document} 

\maketitle 

\thispagestyle{empty} 


\begin{abstract}
Recently, privacy has a growing importance in several domains, especially in street-view images. The conventional way to achieve this is to automatically detect and blur sensitive information from these images. However, the processing cost of blurring increases with the ever growing resolution of images. We propose a system that is cost-effective even after increasing the resolution by a factor of 2.5. The new system utilizes depth data obtained from LiDAR to significantly reduce the search space for detection, thereby reducing the processing cost. Besides this, we test several detectors after reducing the detection space and provide an alternative solution based on state-of-the-art deep learning detectors to the existing HoG-SVM-Deep system that is faster and has a higher performance.
\end{abstract}

\section{Introduction}
\label{sec:intro}

Several street-view services such as Google Street View, Bing Maps Streetside, Mapillary have systematically collected and hosted millions of images. However, the privacy of an individual is vital and hence concealing this information is important for privacy protection. Typically, this is achieved by detecting the privacy-sensitive content and then blurring it. However, this becomes increasingly challenging with growing image resolution. In a commercial setting, it is preferred to exploit ultra-high resolution images which are updated frequently. A higher resolution offers a clearer view of the important content and allows to extract specific details, while the frequent updates provide better tracking of possible changes. However, these benefits come with higher production effort of images that incurs further processing cost for privacy protection. In this research for street-view imaging, since the commercial imaging vendor Cyclomedia transitions from a 100-Megapixel (Mpix) to a 250-Mpix camera system with LiDAR, we present a privacy protection solution, which offers better performance and similar processing cost to the existing system.

Currently, Cyclomedia offers 100-Mpix 360$\degree$ panoramic images (cycloramas), which are taken at a driving interval of every 5 meters. The new system will have 250-Mpix cycloramas along with LiDAR data acquired at the same capturing frequency. However, the current detection algorithm has a quadratic computation cost that makes it infeasible for the processing of the 250-Mpix images. The proposed system exploits the LiDAR to preprocess the 250-Mpix cycloramas such that the search space is largely reduced for the detection algorithm. To account for the quadratic cost of the current detector, we replace it with a Convolutional Neural Network (CNN) detector that has a linear cost for computation. We extensively test several detectors to reach the best trade-off between speed and performance. Finally, the proposed method achieves a higher detection performance and a computation cost similar to the 100-Mpix system.

\section{Related Work}

Various approaches are available for privacy protection in street-view images. The typical way to achieve this is to detect and blur objects in images. One of the first approach for privacy protection explores a two-stage detector \cite{largescaleICCV2009}. The first stage comprises of two detectors running in parallel, one with high recall and another with high precision. This is followed by a post-processing step that converts the output regions in the bounding box to a feature vector that is passed to a neural network. If the output from the neural network exceeds a certain threshold, then it is blurred. Apart from detection and blurring methods, there are only a few approaches that simultaneously detect, remove and inpaint pedestrians \cite{replacePedesICPR2012, removePedesCVPRW2010}. However, inpainting based on the context leads to inaccurate content and is not preferred for commercial applications.

With the advances in deep learning, object detection has improved significantly. Object detection provides a reliable way to localize and detect objects of interest. Several deep learning-based object detection algorithms are available with trade-offs between speed and performance. They are built on top of CNN architectures such as ResNet and its variants \cite{ResNetCVPR2015, Inceptionv4AAAI2017}. Each of the CNN architectures has a meta-architecture on top that completes the object detection framework. Popular meta-architectures include Faster-RCNN, Region-based Fully Convolutional Network (RFCN) and Single Shot Multi-box detector (SSD) \cite{FRCNNNIPS2015, RFCNNIPS2016, SSDECCV2016}. Both Faster-RCNN and RFCN are two-stage detectors. Faster-RCNN has a region proposal network (RPN) that utilizes predefined anchor boxes which are later refined, whereas RFCN has a position-sensitive RoI pooling that encodes class-specific positional information into specific feature maps. SSD is a single-stage detector that directly predicts a bounding box and its score at each spatial position using a small convolution layer. Compared to both Faster-RCNN and RFCN, SSD is faster due to its single-stage approach. However, the two-stage networks offer better performance. In this research, we compare the performance of these meta-architectures on top of different CNN architectures, in order to find a detector that is suited to replace the existing face and license plate detectors.

\begin{figure*}[t]
    \centering
    \includegraphics[width=500pt, height=240pt]{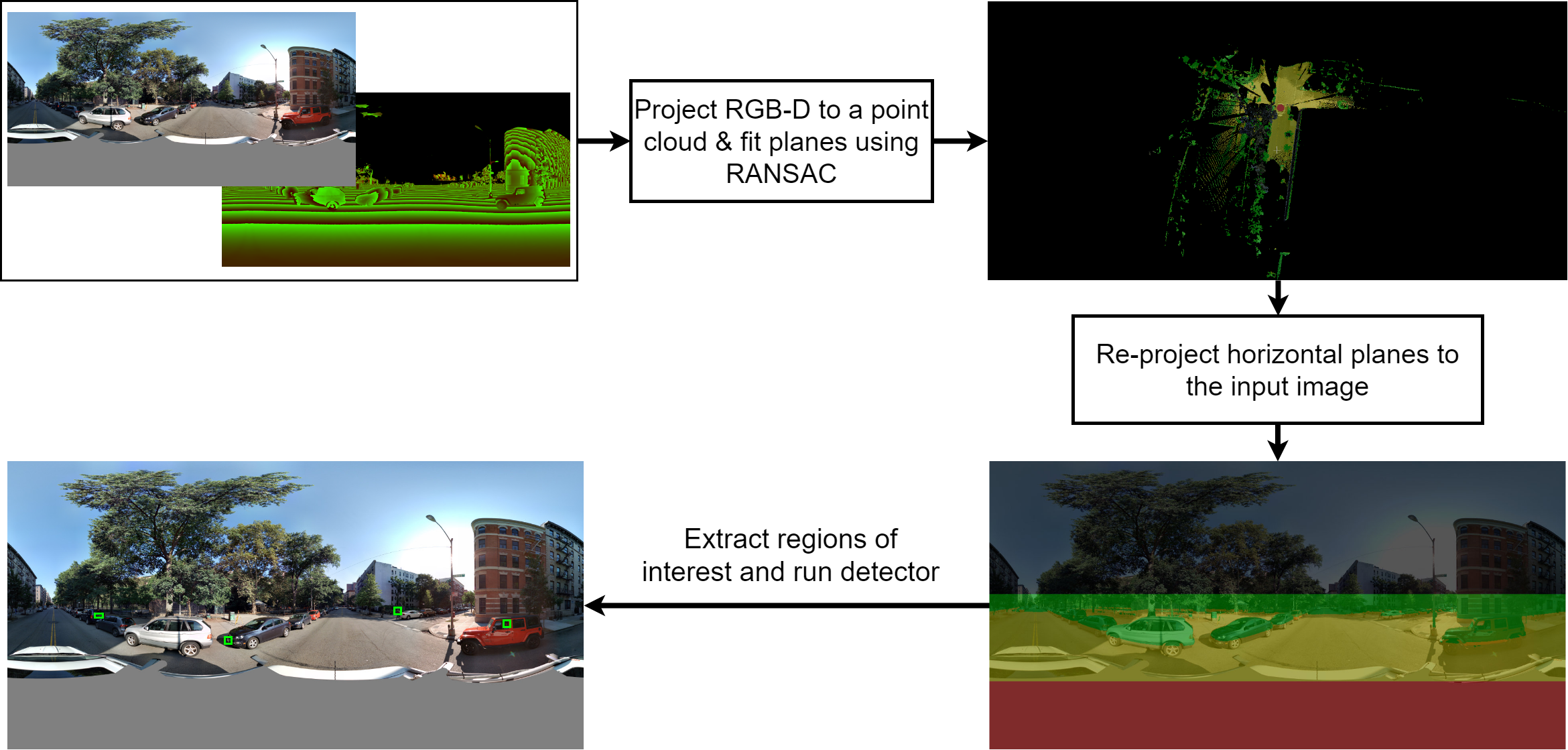}
    \caption{Overview of the new detection system. The input RGB-D image is converted to a point cloud and the planes are extracted using RANSAC. The found horizontal planes are combined and are reprojected to an image plane. The red region indicates the removed areas of the ego-vehicle, the yellow region shows the extracted ground planes and the green region depicts the buffer added to the top boundary of the extracted planes (bottom-right image). The extracted areas (green and yellow regions) are processed by the object detector.}
    \label{fig:Overview}
\end{figure*}

Nowadays, LiDAR-based system are becoming more prevalent in the autonomous driving industry. RGB-D-based object detection has been explored in \cite{RGBDICME2017, RGBDIJRR2017}. However, to the best of our knowledge, this is the first work that utilizes depth obtained from LiDAR to reduce the search space of a detection algorithm. Plane fitting in point clouds using Random Sample Consensus (RANSAC) is well studied and several improvements have been proposed \cite{RANSACACM1981}. In this research, we utilize depth data obtained from LiDAR to create a point cloud which is later used to estimate planes. The estimated planes are further applied to reduce the search space in the cycloramas.

\section{Overview of Architecture}

The final objective of the ideal 250-Mpix system is to have a processing cost (on the cloud) similar to that of the current 100-Mpix system. The current system is deployed on CPU, whereas the proposed system utilizes the GPU architecture. To provide a fair comparison, we compare the processing time with a specific constraint. The constraint is that cost of processing on the GPU is 6 times more expensive than the CPU. Hence, the new system should process a single image in one-sixth of the  current system's processing time. The emphasis is on achieving high recall since privacy protection is crucial. False positives that may lower the precision can be manually removed.

\subsection{Camera system and dataset}
Currently, the cycloramas (360$\degree$ panoramas) are captured at the 100-Mpix resolution. They are processed by the current detection system on CPUs in the cloud. The new camera system captures 250-Mpix cycloramas along with the LiDAR data. The LiDAR data is processed to produce a depth map corresponding to every pixel in the RGB image. The dataset consist of approximately 3000 cycloramas with 50,000 objects (faces and license plates). The images are taken from various parts of the Netherlands, United States and Germany. The dataset is divided in the ratio of 70:30 for training and test sets.

\subsection{Current detection system}
The current system at Cyclomedia is composed of an ensemble of eight cascaded HoG-SVM detectors. Five of the detectors are trained to detect faces, whereas three of them are used for the detection of license plates. Each of them is a HoG feature-based SVM object detector that is stacked three times forming a cascade, followed by a CNN classifier that is used to remove false positives. The detectors try to achieve a high recall at the cost of higher false positives. Each detector is also trained with hard negative mining to reduce false positives. The detectors follow a multi-scale approach for training and inference. Each image is processed over 50 scales to maximize the recall. For simplicity, we refer to this complex pipeline as HoG-SVM-Deep. The disadvantage of this system is that with growing resolution, the number of scales that needs to be scanned for the objects of interest also increases. Empirically, we have found that the cost of processing increases approximately by 6.5 times in the transition from 100-Mpix to the 250-Mpix system.

\subsection{Proposed detection system}
\begin{figure*}[t]
    \centering
    \includegraphics[width=500pt, height=223pt]{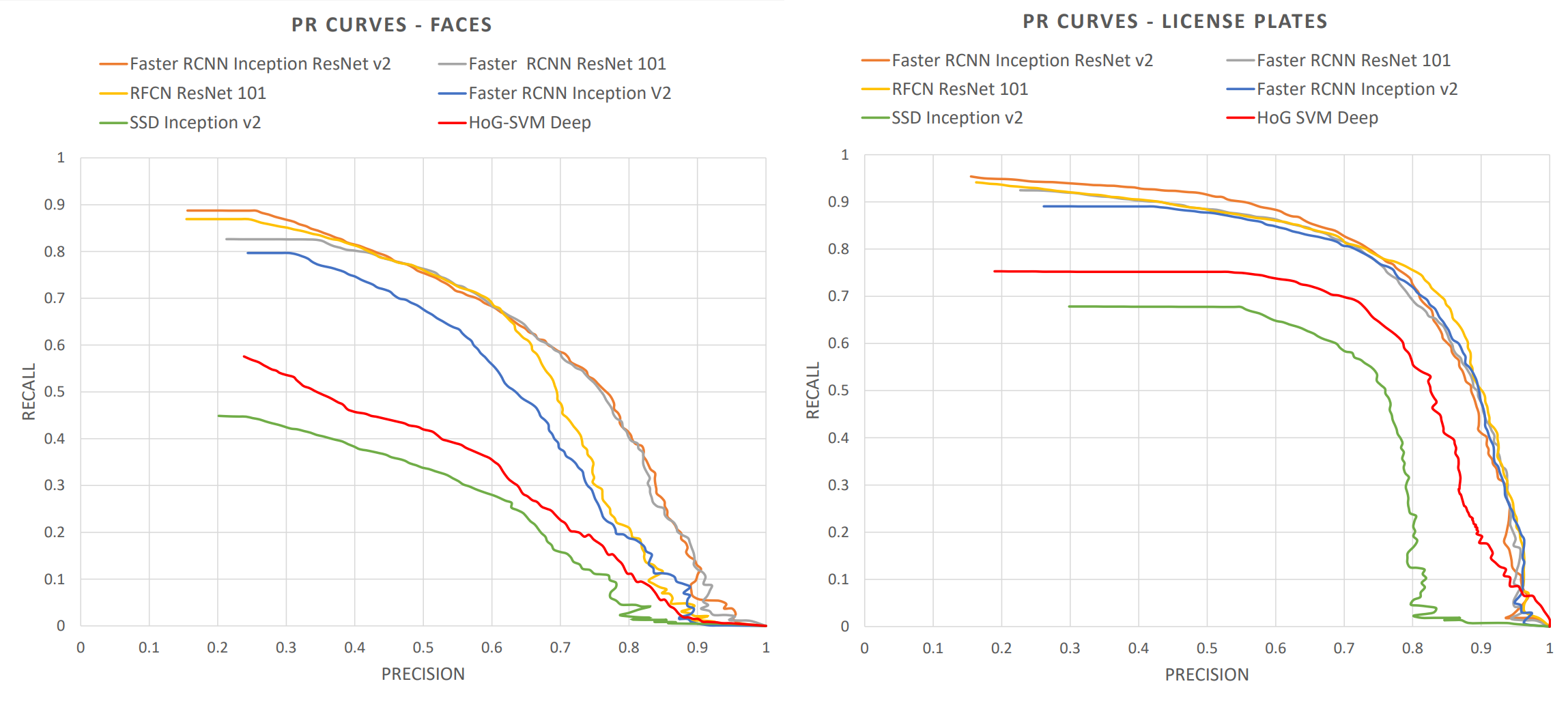}
    \caption{Precision recall curves for different detectors for face and license plate detectors.}
    \label{fig:PR}
\end{figure*}
The proposed system is composed of two parts, a pre-processing step to reduce the search space for the detection algorithm, and a deep learning detector step which offers a good trade-off between speed and performance. An overview of the proposed system is shown in Figure \ref{fig:Overview}. A cyclorama may contain objects of interest at different parts of an image. However, some regions such as sky or the area covered by the ego-vehicle do not inherently contain any objects. In the current system, these regions are removed prior to the detection and only 66\% of a cyclorama is processed by the detector. However, the regions where the object of interest is present are
dependent on the ground plane. We expect objects such as faces and license plates to be present at
most two meters above the ground level. We utilize the depth maps (calibrated and corrected) from the
LIDAR scanners to estimate the ground planes. We first project the depth map to a point cloud, then fit the planes using RANSAC.

\begin{figure*}[t]
    \centering
    \includegraphics[width=500pt, height=84pt]{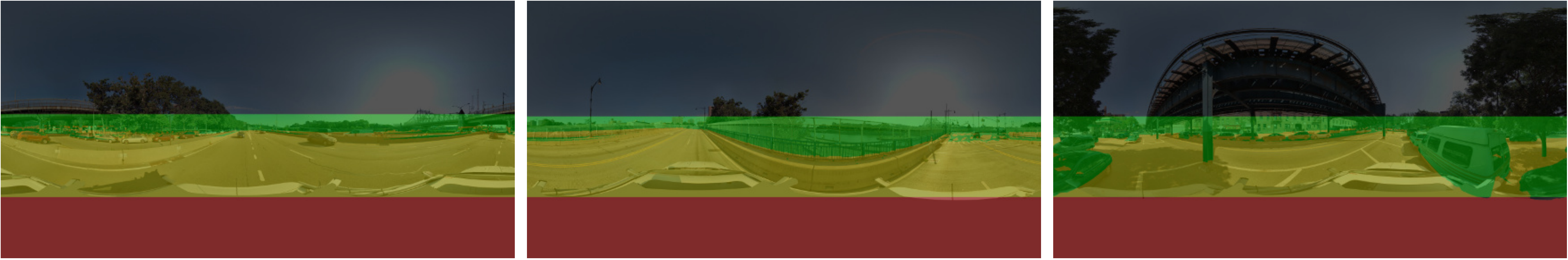}
    \caption{Outputs from plane extraction algorithm. The red and the dark regions of the image are not processed. The bands of yellow and green area is processed by the detector.}
    \label{fig:planes}
\end{figure*}
Projecting a 250-Mpix image to a point cloud and fitting planes are computationally expensive operations. Therefore, the RGB-D image is downsampled prior to projecting to a point cloud. The image is downsampled by a factor of 10 in both $x-$ and $y-$ directions. The planar segmentation results in vertical and horizontal planes. We filter out the vertical planes and retain the horizontal planes, as they may contain people or vehicles. The horizontal surfaces are reprojected into the image plane, to obtain a mask that possibly contains the objects of interest. We select a rectangular region around the boundaries of the mask, with the region size at about one-third of the cyclorama height. 
 
 The region obtained from the mask has high resolution that cannot be directly processed by a CNN. Therefore, it is converted into patches of 1200 $\times$ 600 pixels which is later supplied to the detection algorithm. Due to the dynamic nature of the ground planar segmentation, there is a significant reduction in the area for object detection. We have found that on average only 31\% of the cyclorama needs to be processed by the detection algorithm. This is less than one half of the manually estimated processing area in the current detection system. 

\begin{table}[!ht]
\caption{Table 1: Time vs. Performance of 100-Mpix and 250-Mpix camera systems on face and license plate detectors.}
\label{tab:performance}
\begin{center}       
\begin{tabular}{|c|c|c|c|c|} 
\hline

\multirow{2}*{\textbf{Detector}} & \multicolumn{2}{c|}{\textbf{Time (sec)}} & \multicolumn{2}{c|}{\textbf{Max Recall}} \\

 & current & new & Face & L.P. \\ \hline

SSD Inception v2 & 16.50 & 16.98 & 44.8 & 67.8 \\ \hline
FRCNN Inception v2 & 21.15 & 21.54 & 79.6 & 89.0 \\ \hline
RFCN ResNet-101 & 44.64 & 45.94 & 86.9 & 94.1 \\ \hline
FRCNN ResNet-101 & 86.54 & 89.06 & 82.6 & 92.4 \\ \hline
FRCNN In. ResNet v2 & 378.68 & 389.62 & 88.7 & 95.3 \\ \hline
HoG-SVM-Deep & 215 & 879.12 & 57.8 & 75.2\\ \hline

\end{tabular}
\end{center}
\end{table} 

\section{Experiments and Results}
We have experimented with several deep learning detectors and compared them with the currently deployed existing HoG-SVM-Deep system. We have tested deep learning detectors with three meta-architectures, namely SSD, RFCN, and Faster-RCNN (abbreviated as FRCNN in the table) with Inception v2, ResNet-101 and Inception ResNet v2, as the backbone architecture.  We train on the above-mentioned dataset with a model pretrained on the COCO dataset. We train each network for 200,000 iterations with a batch size of 4 for all networks except SSD Inception v2. The SSD Inception v2 network uses a batch size of 24 with a learning rate of $1^{-05}$. The learning rates for the rest of the networks are set to $1^{-06}$.  The computational costs of the different detectors are presented in Table~\ref{tab:performance}. Note that the reported times are GPU execution times for the deep learning-based detectors, whereas the HoG-SVM-Deep detector uses a CPU. The HoG-SVM-Deep system is executed on the CPU as it is cheaper to process on the cloud services, whereas the deep learning-based detectors are more efficiently processed on the GPU. HoG-SVM-Deep is executed on a Xeon E5-2673 v3 CPU, whereas the rest of the detectors are executed on a Tesla K80 GPU.

The plane extraction is performed using RANSAC. Three points are selected and the corresponding parameters are computed. Depending on the given threshold, the outliers are removed and a plane is fitted. This procedure is repeated until a plane with a maximum number of points is obtained. Once a plane is estimated, the points corresponding to that plane are removed and a new plane is computed. This process is repeated until the top-10 planes are found. Usually, the largest plane is the ground plane. However, we combine the largest 10 planes in the output image after reprojection of the point cloud, to maximize the area of ground surface. We have tuned the RANSAC-based plane fitting to our dataset. In our experiments, we have found that setting the distance threshold (distance at which a point is considered as an inlier for the model) to 0.5 meters results in the best estimation of planes. After the planes are estimated and reprojected to the image plane, we add a small buffer of 350 pixels (green region in Figure \ref{fig:planes}) to the top boundary of the extracted plane (yellow region in Figure \ref{fig:planes}) on the reprojected image. We have observed that such a small buffer is needed to account for people standing on the ground farther away from the image. It was empirically found that adding 350 pixels can account for all the objects of interest present in our dataset.

After the planar segmentation and subsequent patching, each patch of 1200 $\times$ 600 pixels is fed to the detector. The detector performance on our dataset is presented in Figure~\ref{fig:PR} and Table~\ref{tab:performance}.  All the networks except SSD Inception v2, offer better performance than HoG-SVM-Deep for both face and license plate detection. Only Faster-RCNN with Inception ResNet v2 has higher computation cost than HoG-SVM-Deep, but offers significantly higher performance. Comparing the execution times of the current and new system of all the detectors, it can be seen that they yield very similar computation time. This is due to the search-space reduction achieved using depth data on the new camera system. The processing cost of executing the current system on the CPU is 215 seconds. Following the constraint mentioned in the previous section, we need a system that utilizes one-sixth of the current system (~36 seconds). From Table \ref{tab:performance}, only Faster-RCNN with Inception v2 satisfies that requirement. However, it offers much higher performance than the current system. On the other hand, RFCN with ResNet-101 have higher performance at the expense of 28\% longer processing time. Both options form feasible solutions for the existing system.  

\subsection{Limitations}

We have utilized the RANSAC plane extraction algorithm, as it is effective and fast. However, this may be challenging in locations where the surfaces are not flat. The underlying assumption that objects such as faces can be found only on the ground plane have exceptions in scenarios where people are present on top of a building. We found that such cases occur once in every 50,000 images. In such scenarios, the objects are manually blurred.

\section{Conclusion}
We have presented a framework that is an alternative to the existing system used at Cyclomedia for privacy protection in street-view images. The current system utilizes HoG-SVM-Deep to process a 100-Mpix image. We propose a new detection system for the 250-Mpix images that utilizes the depth data obtained from LiDAR to reduce the search space. The reduction results in processing only 31\% of the 250-Mpix cyclorama. We obtain detectors that are suited for our problem, namely Faster-RCNN with Inception ResNet v2 and RFCN with ResNet-101. Both of these detectors offer a higher recall than the existing system for detecting both face and license plates, providing 37-50\% and 18-25\% improvement in recall on face and license plate detection. On the other hand, the proposed system using the new detectors results in processing costs that are comparable to the existing 100-Mpix camera system.




\end{document}